\pdfoutput=1

\documentclass[11pt]{article}

\usepackage[]{EMNLP2022}

\usepackage{times}
\usepackage{latexsym}

\usepackage[T1]{fontenc}

\usepackage[utf8]{inputenc}

\usepackage{microtype}

\usepackage{inconsolata}

\usepackage{graphicx}
\usepackage{booktabs,multirow}
\usepackage{caption}
\usepackage[whole]{bxcjkjatype}
\usepackage{bbding}
\usepackage{pifont}
\usepackage{subcaption}
\usepackage{lingmacros}
\usepackage{tikz-dependency}
\usepackage{pifont}
\usepackage{amssymb,mathtools}

\usepackage{color}
\usepackage{caption}
\usetikzlibrary{positioning}  
\usepackage{bm}
\usepackage{float}
\usepackage{tikz}
\usepackage{caption}
\usepackage{fontawesome}

\title{Context Limitations Make Neural Language Models More Human-Like}

\author{Tatsuki Kuribayashi$^{1,2}$ $\;\;\;$ Yohei Oseki$^{3,4}$ $\;\;\;$ Ana Brassard$^{1,4}$  $\;\;\;$ Kentaro Inui$^{1,4}$ \\
 $^1$Tohoku University $\;$ 
 $^2$Langsmith Inc. $\;$ 
 $^3$University of Tokyo $\;$ 
 $^4$RIKEN \\
\texttt{\{kuribayashi, inui\}@tohoku.ac.jp } \\
\texttt{oseki@g.ecc.u-tokyo.ac.jp} $\;\;\;$ \texttt{ana.brassard@riken.jp }
}

\begin{document}
\maketitle

\begin{abstract}
Language models (LMs) have been used in cognitive modeling as well as engineering studies---they compute information-theoretic complexity metrics that simulate humans' cognitive load during reading.
This study highlights a limitation of modern neural LMs as the model of choice for this purpose: there is a discrepancy between their context access capacities and that of humans.
Our results showed that \textit{constraining} the LMs' context access improved their simulation of human reading behavior.
We also showed that LM-human gaps in context access were associated with specific syntactic constructions; incorporating syntactic biases into LMs' context access might enhance their cognitive plausibility.\footnote{Our codes are available at \faGithub\ \url{https://github.com/kuribayashi4/context_limitation_cognitive_modeling}}
\end{abstract}

\section{Introduction}
\label{sec:intro}

In computational psycholinguistics, human reading behavior has been compared with various complexity metrics to understand human sentence processing~\cite{Crocker2010-cp}.
Having historically started from simple measures such as word length, surprisal ($-\log p(\mathrm{word}|\mathrm{context})$) computed by language models (LMs) has become a common choice~\cite{Levy2008Expectation-basedComprehension,Smith2013TheLogarithmic}. 
On top of this, the next question arises---which model implementation and/or algorithm can compute surprisal that successfully simulates human behavior?
In this line of research, modern neural LMs such as Transformer~\cite{vaswani2017} have been analyzed with respect to their cognitive plausibility~\cite{Wilcox2020OnBehavior,Merkx2020ComparingData,kuribayashi-etal-2021-lower}.

\begin{figure}[t]
    \centering
      \includegraphics[width=\linewidth]{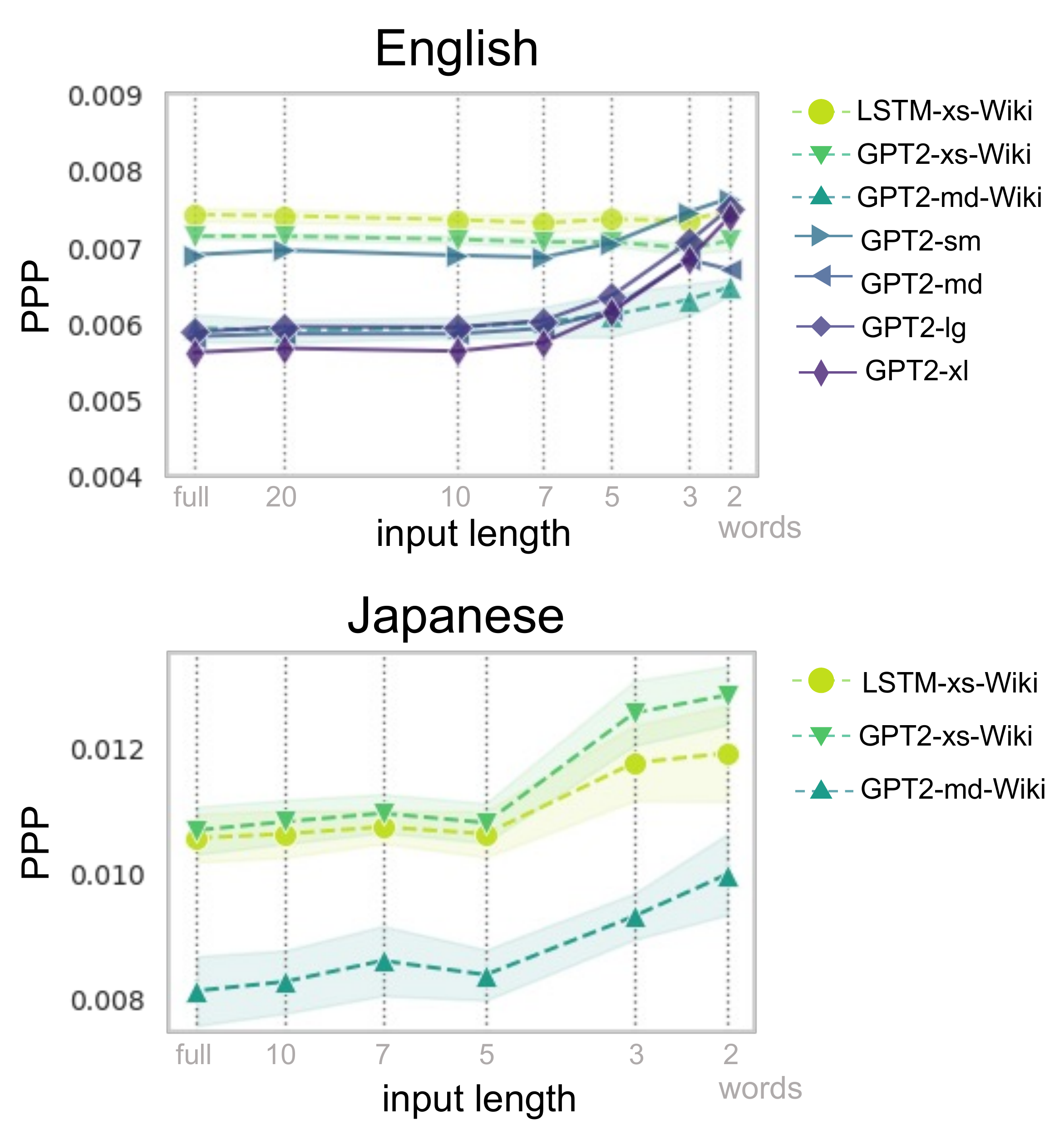}
      \caption{Relationship between psychometric predictive power (PPP) of language models (LMs) and their context access constraints.
      LMs with less context access better simulate human reading behavior (higher PPP).
      The marker color/shape indicates LM settings; colored areas present one standard deviation of PPP.
      }
      \setlength{\belowcaptionskip}{-100pt}
      \label{fig:ppp}
\end{figure}

\begin{table*}[t]
    \centering
\centering
\renewcommand{\arraystretch}{0.9}
{\small
\begin{tabular}{p{5.5cm}p{1.6cm}p{1.5cm}p{1.5cm}p{2.3cm}p{1cm}} \toprule
 Neural model architecture & Data size & Training steps & Tokenization & Syntactic supervision & Context length \\
\cmidrule(lr){1-1} \cmidrule(lr){2-2} \cmidrule(lr){3-3}  \cmidrule(lr){4-4}  \cmidrule(lr){5-5}  \cmidrule(lr){6-6}
\citet{Goodkind2018PredictiveQuality,Aurnhammer2019-ib,Wilcox2020OnBehavior,Hao2020-wo,Merkx2020ComparingData,Oh2021-ln,kuribayashi-etal-2021-lower} & \citet{Wilcox2020OnBehavior,kuribayashi-etal-2021-lower} & \citet{kuribayashi-etal-2021-lower} & \citet{Wilcox2020OnBehavior,Oh2021-ln} & \citet{Hale2018FindingSearch,Yoshida2021-rc} & \textbf{This work} \\
\bottomrule
\end{tabular}
}

\caption{Related studies exploring psychometric predictive power of neural models while \textit{separately} controlling a specific factor of their configuration.
}
\label{tbl:related_study}
\end{table*}

Despite their use in cognitive modeling, such modern LM architectures (e.g., self-attention)  are, arguably, an unnatural choice when it comes to human cognitive constraints; modern LM architectures assume powerful, parallel access to a vast number of context tokens, while humans might have limited and selective context access~\cite{hawkins1994performance,Gibson1998LinguisticDependencies,Gibson2000TheComplexity,Lewis2006-qs}.
Searching for a computational model that better simulates human sentence processing than previously examined ones, we hypothesized that introducing such context limitations can improve LMs' estimation of human cognitive load.

Specifically, as a starting point, we applied an n-gram-ification trick to neural LMs mimicking loading for long context access (\textit{locality effects}) and compared their surprisal with human reading behavior data.
Despite the simple context limitation design, our experiments with 280 settings (40 LM settings$\times$7 noise patterns) showed that the advantage of a shorter context was consistent among LMs and typologically different languages (Figure~\ref{fig:ppp}).
This showed that \textit{constraining} the modern LM's context access is key to increasing their similarity to the model of human reading.

Furthermore, expecting that humans' context limitations might be more complex than simple distance-based erasure, we conducted exploratory analysis of in which constructions longer/shorter contexts were beneficial.
We found that the context limitation (dis)advantages were allocated in specific syntactic constructions, suggesting that, to build more cognitively plausible LMs, adding syntactic biases in their context access could be beneficial. 
From a psycholinguistic view, our results empirically highlight the memory account of human sentence processing during naturalistic reading~\cite{Futrell2020Lossy-ContextProcessing}.

\section{Background}

\subsection{Human sentence processing}
Humans incrementally process text and exhibit different processing costs (e.g., reading times) for different tokens. 
Psycholinguistic theories on such processing costs are divided between expectation-based and memory-based perspectives.

\textit{Expectation-based} theories claim that humans predict upcoming words during incremental sentence processing~\cite{Clark2013-xs}.
Recent studies have extensively analyzed this expectation-based aspect by comparing surprisal, $-\log p(\mathrm{word}|\mathrm{context})$, to human reading behavior~\cite{hale-2001-probabilistic,Levy2008Expectation-basedComprehension,Wilcox2020OnBehavior}.

On the other hand, \textit{memory-based} theories have asserted that human sentence processing is constrained by a limited cognitive resource~\cite{Gibson2000TheComplexity,Lewis2005-hp,Lewis2006-qs}. 
Cross-linguistic studies have reported that different languages incur different memory decay during reading~\cite{Konieczny2000LocalityComplexity,Vasishth2010-ji,Husain2014-zr,Frank2016-so}.
Notably, memory efficiency is also considered in exploring the design (e.g., word order) and evolution of natural language~\cite{Greenberg1963-rm,Chomsky2005-he,Gibson2019-oe,Hahn2020-sb,Hahn2020-dq}.

Recently, \citet{Futrell2017noisy} and \citet{Futrell2020Lossy-ContextProcessing} have proposed integrating the two theories through the concept of \emph{lossy-context surprisal}---next-word probabilities calculated with \emph{noisy} context should better predict human reading behavior than with complete context.
These studies have focused on its theoretical aspects and explaining a specific phenomenon (e.g., verb forgetting); on top of this, our study demonstrates the theory's broad benefit in modeling naturalistic reading data.

Notably, such a simulation of human cognitive load also contributes to achieving text readability assessment~\cite{Ambati2016}.
Furthermore, human-like agents are necessary in \textit{in silico} simulation studies on language evolution~\cite{Galke2021,Rita2021,ueda2021}.

\subsection{Cognitive plausibility of LMs}
Surprisal from certain LMs could predict human reading behavior well; thus, what \emph{type} of LM does better simulate human reading behavior?
LM-based analyses have typically explored inductive biases, such as LM architecture (Table~\ref{tbl:related_study}).
We focus on context limitation as an alternative factor.

Studies comparing the cognitive plausibility of LM architectures also addressed, albeit implicitly, context access abilities~\cite{Aurnhammer2019-ib,Merkx2020ComparingData}.
For example, simple recurrent neural networks assume relatively weak context access, whereas Transformer LMs~\cite{vaswani2017} assume stronger access than those~\cite{Michaelov2021-cx,Merkx2020ComparingData}.
In addition, studies contrasted count-based n-gram LMs and neural LMs~\cite{Wilcox2020OnBehavior,Hao2020-wo,Goodkind2018PredictiveQuality}; however, (i) count-based versus neural-based estimation and (ii) partial versus full context access were not distinguished.
By contrast, we fixed the architectures and investigated the exact effect of context access with input deletion.

\section{Methods}
\label{sec:method}

We investigate how human-like neural LMs become with more or less context at their input.
Specifically, we measured the psychometric predictive power (PPP) of lossy-context LM surprisal for gaze duration modeling.
In the following sections, we describe each measure in detail.%

\subsection{Psychometric predictive power}
\label{subsubsec:psycho_pw}
In this study, the cognitive plausibility of a model $\theta$ is measured via the similarity between its surprisal and human gaze duration across words based on surprisal theory~\cite{Smith2013TheLogarithmic,Levy2008Expectation-basedComprehension}.
Here, the surprisal of a word computed by a model $\theta$, $-\log p_\theta(\mathrm{word}|\mathrm{context})$, is compared with the corresponding word's gaze duration.

Specifically, we measured the \textbf{psychometric predictive power (PPP)} of surprisal values by fitting two nested linear mixed-effects regression models that predict gaze duration, one with surprisal features and the other without. 
Here, the per-token difference in their log-likelihoods ($\Delta$LogLik; LogLik with surprisal minus LogLik without surprisal) denotes PPP, following~\citet{Goodkind2018PredictiveQuality}.
The larger the PPP ($\Delta$LogLik), the more useful the surprisal for modeling gaze duration, i.e., the model computes surprisal well correlating with human behavior. 
See Appendix~\ref{app:feature} for detailed features used in regression modeling.

\subsection{Lossy-context surprisal}
\label{subsubsec:lossy_context_surprisal}

Instead of the full-context surprisal, we investigate the PPP of surprisal conditioned by limited context $-\log p_{\theta}(\mathrm{word}|\mathrm{lossy\_context})$ to explore the cognitive plausibility of context-limited LMs~\cite{Futrell2020Lossy-ContextProcessing}. 
The \textbf{lossy-context surprisal} of the symbol $w_i$ given its preceding context $c_{<i}=[w_0, \cdots, w_{i-1}]$ is defined as follows:

\vspace{-0.4cm}
\begin{align}
\nonumber
    &I_\mathrm{lossy}(w_i, c_{<i}, f) \\ 
    \label{eq:surprisal}
    &= -\log p_{\theta}(w_{i}| \mathrm{\texttt{<s>}} \circ f([w_0, \cdots, w_{i-1}])) \:\:,
\end{align}

\noindent
where $\theta$ denotes left-to-right LMs,  \texttt{<s>} denotes the beginning of a sequence, $\circ$ is a concatenation function, and $f$ represents a noise function.
The noise function controls the LMs' access to contextual information by deleting the input of LMs with a particular pattern.
For example, if $f$ is leaving only the last two symbols, $I_\mathrm{lossy}$ corresponds to surprisal from 3-gram LMs, and if $f$ is an identity function, $I_\mathrm{lossy}$ corresponds to unmodified surprisal.

Gaze duration is typically annotated in larger spans such as words, while LMs' input is at the smaller levels (i.e., subwords).
The lossy-context surprisal of a span $s = [w_l,w_{l+1},\cdots,w_{m}]$ ($0\leqq l < m$)  was calculated as the cumulative surprisal of its constituent subwords:

\vspace{-0.3cm}
\begin{equation}
    I_\mathrm{lossy}(s, c_{<l}, f) = \sum_{j=l}^{m} I_\mathrm{lossy}(w_j, c_{<j}, f) \:\:.
    \label{eq:lossy_surprisal_segment}
\end{equation}

\paragraph{$N$-gram surprisal.}
As a starting point, based on the assumption about human working memory that distant context is hard to access~\cite{Lewis2006-qs}, we explored surprisal given by LMs conditioned on $n-1$ preceding words (not subwords);
henceforth, this surprisal is referred to as $n$-gram surprisal (a special case of lossy-context surprisal).
In Appendix~\ref{app:probabilistic_noise}, we also explored a probabilistic version of the noise inspired by~\citet{Futrell2020Lossy-ContextProcessing}, yielding consistent conclusions with our experiments using $n$-gram surprisal.

\subsection{Gaze duration}
\label{subsec:eye_track_data}

Gaze duration data were modeled by lossy-context surprisal.
To explore the cross-linguistic consistency of our results, we used two typologically different languages, English and Japanese; their difference is introduced in the later paragraph.

\begin{table}[t]
    \centering
\centering
\begin{tabular}{lrrr} \toprule
Data & sentLen & contextLen & wordLen \\ %
\cmidrule(lr){1-1} \cmidrule(lr){2-2} \cmidrule(lr){3-3}  \cmidrule(lr){4-4}
DC &  17.8$\pm$11.7 & 13.0$\pm$10.1 & 1.3$\pm$0.7 \\
BE & 7.9$\pm$5.3 & 6.0$\pm$4.6 & 3.4$\pm$2.3 \\ 
\bottomrule
\end{tabular}

\caption{Statistics of all the sentences in each corpus. The values present mean$\pm$standard deviation; sentLen denotes the number of words in a sentence, contextLen denotes the number of preceding words within the same sentence for each word, and wordLen denotes the number of subwords in each word.}
\label{tbl:stats}
\end{table}

\paragraph{Data.}
For English, we used the Dundee Corpus (DC)~\cite{kennedy2003dundee}.
As its Japanese counterpart, we used BCCWJ-EyeTrack (BE)~\cite{Asahara2016Reading-TimeJapanese}.
In both corpora, first-pass gaze duration information was used.
The average sentence/context lengths are shown in Table~\ref{tbl:stats}.
Note that while the English gaze duration annotation is typically attached to space-separated words, Japanese gaze duration annotation is attached to each phrasal unit (\textit{bunsetsu}; henceforth, ``word''); Japanese ``words'' contain more subwords than English words.
Following~\citet{Goodkind2018PredictiveQuality}, we excluded outliers such as words with special characters (details in Appendix~\ref{app:data}).
We used 212,649 data points from DC and 9,217 from BE.

\paragraph{Cross-linguistic analysis.}
English and Japanese sentence structures differ in their branching directions; while English word order (SVO) has mixed directionalities of head-initial and head-final dependencies, Japanese word order (SOV) strongly prefers head-final, left-branching constructions.
The dependency structures of the sentence ``\textit{the dog wagging its tail eats fish on the desk.}'' in English and Japanese are contrasted below:\footnote{For simplicity, some functional words (e.g., ``the,'' ``を'') are merged into a single node.}

\vspace{0.5cm}
{\small

     \begin{dependency}[theme = simple, arc angle=30]
     \begin{deptext}[column sep=0.2em]
     (1) \& The dog \& wagging \& its tail \& ate \& fish \& on the desk.  \\
    \end{deptext}

    \depedge[black]{2}{3}{}
    \depedge[black]{3}{4}{}
    \depedge[blue]{5}{2}{}
    \depedge[black]{5}{6}{}
    \depedge[black]{5}{7}{}
  \end{dependency}
  
  \begin{dependency}[theme = simple, arc angle=30]

    \begin{deptext}[column sep=0em]
      (2) \& 尻尾を \& 振る \& 犬が \& 机の \& 上で \& 魚を \& 食べた。  \\
      \& tail \& wagging \& dog \& on \& desk \& fish \& ate \\
    \end{deptext}

    \depedge[blue]{3}{2}{}
    \depedge[blue]{4}{3}{}
    \depedge[blue]{8}{4}{}
    \depedge[blue]{6}{5}{}
    \depedge[blue]{8}{6}{}
    \depedge[blue]{8}{7}{}
  \end{dependency}
 }

Such an  asymmetry of structure has been reported to incur different memory biases for sentence processing~\cite{Konieczny2000LocalityComplexity,Nakatani2008DistinguishingJapanese,Vasishth2010-ji,Futrell2020Lossy-ContextProcessing}.
Thus, we included typologically different languages in our experiments.

\section{Language models}
\label{sec:lm}

We used two types of neural LMs for lossy-context (n-gram) surprisal computation: (i) Wiki-LMs and  (ii) pretrained OpenAI GPT-2s~\cite{Radford_undated-nn}.
Their hyperparameters are shown in Appendix~\ref{app:hyper}.
Notably, using \textit{neural} LMs makes the comparison of long-context and short-context LMs computationally tractable.\footnote{For example, if we use \textit{count-based} LMs~\cite{heafield-etal-2013-scalable}, even a single 5-gram Japanese LM took 27GB in model size.}

\subsection{Wiki-LMs}
\label{subsec:wiki-lms}

\paragraph{Model settings.}
We used three variants of unidirectional neural LM architectures: LSTM-xs-Wiki (27M parameters)~\cite{Hochreiter1997LongMemory}, GPT2-xs-Wiki (29M), and GPT2-md-Wiki (335M)~\cite{vaswani2017}.
We trained each LM with three different random seeds using the \texttt{Fairseq} toolkit~\cite{ott-etal-2019-fairseq}.

In both English and Japanese settings, the input is split into subwords with byte-pair encoding~\cite{Sennrich2016NeuralUnits}.\footnote{Note that there is some debate on the cognitive plausibility of subwords~\cite{Oh2021-ln,Anonymous2022-ui}; we consider this issue to be out of the scope of this study.}
Specifically, for the Japanese data, we adopted two-stage segmentation to ensure that multiple subwords compose a Japanese word defined in a commonly used corpus (e.g., BE).\footnote{This procedure is common practice in Japanese NLP. See \url{https://github.com/himkt/awesome-bert-japanese}. We used Mecab~\cite{kudo2006mecab} with a unidic dictionary (\url{https://unidic.ninjal.ac.jp/}) for morphological analysis.}
That is, text was segmented in advance into morphemes~\cite{Maekawa2014-ze}, and then a subword tokenizer was applied to the morpheme-separated texts.
Details are in Appendix~\ref{app:hyper}.

\begin{table}[t]
    \centering
    \small
    \begin{tabular}{p{7cm}} \toprule
     ... \texttt{<b>} \textcolor{orange}{\_was \_also \_the \_first \_hotel \_in \_Westchester \_County .} \texttt{<b>} \textcolor{blue}{\_on \_4 \_March \_1990 \_with \_a \_concert \_performed \_by \_Ell a \_Fitzgerald \_at \_the \_Royal \_Albert \_Hal} \texttt{<b>} \textcolor{red}{\_the \_Har row} \texttt{<b>} \textcolor{magenta}{\_On \_the \_night \_of \_the \_31 \_May \_/ \_1 \_June \_1941 \_he} \texttt{<b>} ... \\
    \bottomrule
    \end{tabular}
\caption{An example of the modified training data, where sub-sequences (with the same color) sampled from the original corpus were randomly patched. The special token (\texttt{<b>}) indicates the break of contextual dependence between before and after.}
\label{chp5:tbl:example_data}
\end{table}

\begin{table*}[t]
    \centering
\centering
\renewcommand{\arraystretch}{0.8}
\begin{tabular}{clrrrrrrrr} \toprule
&& \multicolumn{7}{c}{Input length} &  \\ %
\cmidrule(lr){3-9}
 Lang. & Model & $\mathrm{full}$ & 20 & 10 & 7 & 5 & 3 & 2 & $\Delta$ \\ 
\cmidrule(lr){1-1} \cmidrule(lr){2-2} \cmidrule(lr){3-3} \cmidrule(lr){4-4} \cmidrule(lr){5-5} \cmidrule(lr){6-6} \cmidrule(lr){7-7} \cmidrule(lr){8-8} \cmidrule(lr){9-9} \cmidrule(lr){10-10}
\multirow{6}{*}{En} & GPT2-xl  & $5.6$ & $5.7$ & $5.6$ & $5.8$ & $6.2$ & $6.8$ & $\mathbf{7.4}^\dagger$ & $1.8$ \\
& GPT2-lg  & $5.9$ & $6.0$ & $6.0$ & $6.0$ & $6.4$ & $7.1$ & $\mathbf{7.5}^\dagger$ & $1.6$ \\
& GPT2-md  & $5.8$ & $5.9$ & $5.9$ & $5.9$ & $6.2$ & $\mathbf{6.8}$& $6.7^\dagger$ & $0.9$ \\
& GPT2-sm & $6.9$ &	$7.0$ &	$6.9$ &	$6.9$ &	$7.1$ &	$7.5$ & $\mathbf{7.6}^\dagger$  & $0.7$ \\
& GPT2-md-Wiki  & $5.9$  &	$5.9$ &	$5.9$ &	$6.0$ &	$6.1$ &	$6.3$ & $\mathbf{6.5}^\dagger$ & $0.6$ \\
& GPT2-xs-Wiki  & $\mathbf{7.1}$ &$\mathbf{7.1}$ &	$\mathbf{7.1}$ &	$\mathbf{7.1}$ &	$\mathbf{7.1}$ & $7.0$ &	$\mathbf{7.1}$  & $0.0$ \\
& LSTM-xs-Wiki & $7.4$ &	$7.4$ &	$7.4$ &	$7.3$ &	$7.4$  & $7.4$ & $\mathbf{7.5}$ & $0.1$ \\
\cmidrule(lr){1-1} \cmidrule(lr){2-2} \cmidrule(lr){3-3} \cmidrule(lr){4-4} \cmidrule(lr){5-5} \cmidrule(lr){6-6} \cmidrule(lr){7-7} \cmidrule(lr){8-8} \cmidrule(lr){9-9} \cmidrule(lr){10-10}
\multirow{3}{*}{Ja} & GPT2-md-Wiki & $8.1$ & $8.1$ &	$8.3$ &	$8.6$ &	$8.4$ &	$9.3$ & $\mathbf{10.0}^\dagger$ & $1.9$  \\
& GPT2-xs-Wiki  & $10.7$ &	$10.7$ &	$10.8$ &	$11.0$ &	$10.8$ &	$12.5$ & $\mathbf{12.8}^\dagger$ & $2.1$\\
& LSTM-xs-Wiki  & $10.6$ &	$10.6$ &	$10.6$ &	$10.7$ &	$10.6$ &	$11.8$ & $\mathbf{11.9}^\dagger$ & $1.3$\\
\bottomrule
\end{tabular}
\caption{Average PPP of $n$-gram surprisal; for example, the input length of 2 corresponds to the PPP of surprisal computed by the neural LMs that take only the 2-gram context as input. For readability, values are multiplied by 1000. The 2-gram PPP with $\dagger$ is significantly higher than its corresponding full-context PPP. The $\Delta$ column shows the PPP gain from the $\mathrm{full}$ context to 2-gram context surprisal in each LM setting.}
\label{tbl:n-gram_lossy}
\end{table*}

\paragraph{Training data.}
For English, the training data were approximately 4M sentences from the WikiText-103 dataset~\cite{Merity2016-rb}, and for Japanese, the data were 4M sentences from Wikipedia and news articles (approximately 0.5GB data size in both English and Japanese).
The sentence order was shuffled, and duplicated sentences were excluded.

\paragraph{Mitigating training-inference mismatches.}
During $n$-gram surprisal computation, LMs must predict the upcoming words with limited context from the middle of a sentence, while such a prediction is rarely enforced during ordinal document-level training.
Such a training-inference mismatch could lead to confusion on whether our results stem from the LM-human gap or biases from the training/inference mismatch.

To handle such a potential mismatch, we modified the LM training data to make the language modeling task more like n-gram one.
Specifically, we randomly split original sentences into smaller chunks of various lengths and randomly patched them by inserting a special token \texttt{<b>} in between the chunks (see Appendix~\ref{app:settings_modify} for the detailed process).
Table~\ref{chp5:tbl:example_data} shows an example.
In this modified corpus, LMs must predict upcoming words by severely limited usable context especially in the data points immediately after the special tokens.
When computing $n$-gram surprisal, the \texttt{<b>} token is set instead of \texttt{<s>} in Eq.~\ref{eq:surprisal}.
Note that this modification does not change the total corpus size.

We trained the Wiki-LMs using this modified data.
In Section~\ref{subsec:main_results}, we ablated the effect of this training modification and showed that such careful training makes the short-context advantage clearer.

\subsection{Pretrained GPT-2s}
\label{subsec:gpt2}

To investigate large-scale LMs typically developed in NLP, we additionally used four variants of pretrained English OpenAI GPT-2s~\cite{Radford_undated-nn}: GPT2-sm (117M params.), GPT2-md (345M), GPT2-lg (774M), and GPT2-xl (1558M).
The input was split into subwords by their pretrained tokenizer with a vocabulary size of 50,257.
The training data were 40GB of web texts.
The potential training-inference mismatch is not handled in the GPT-2 experiments due to the high re-training cost; this point is partially addressed in Section~\ref{subsec:main_results}.
Note that we did not use Japanese versions of pretrained GPT-2s since available models\footnote{\url{https://huggingface.co/rinna/japanese-gpt2-small}} have a tokenizer that is inconsistent with the BE annotation; 16.4\% of word boundaries in the BE were not separated by their pretrained tokenizer.

\section{Experiments}
\label{sec:exp1}

Our experiments demonstrate how limiting context access improved the PPP in LMs, i.e., their surprisal becomes a more effective predictor for human gaze duration (Section~\ref{subsec:main_results}).
As described in Section~\ref{subsubsec:lossy_context_surprisal}, we applied distance-based noise to the input (i.e., computing n-gram surprisal).
A potential training-inference mismatch bias is handled (Section~\ref{subsec:train_settings}).
Furthermore, we explored the connection to existing studies (Section~\ref{subsubsec:perplexity}).

\paragraph{Settings.}
We measured the PPP of 40 LMs (\{3 Wiki-LMs\}$\times$\{3 seeds\}$\times$\{2 languages\}$\times$\{2 training settings\}+\{4 OpenAI GPT-2s\}) with seven noise patterns.
Specifically, we explored the $\{2, 3, 5, 7, 10, 20\}$-grams and $\mathrm{full}$ settings for each LM, where $\mathrm{full}$ refers to using the entire sentence ($w_0$ to $w_{i-1}$) as context.
Note that only intra-sentential context is used as the main focus of this study is sentence-level syntactic processing.

\begin{figure}[t]
    \centering
    \includegraphics[width=\linewidth]{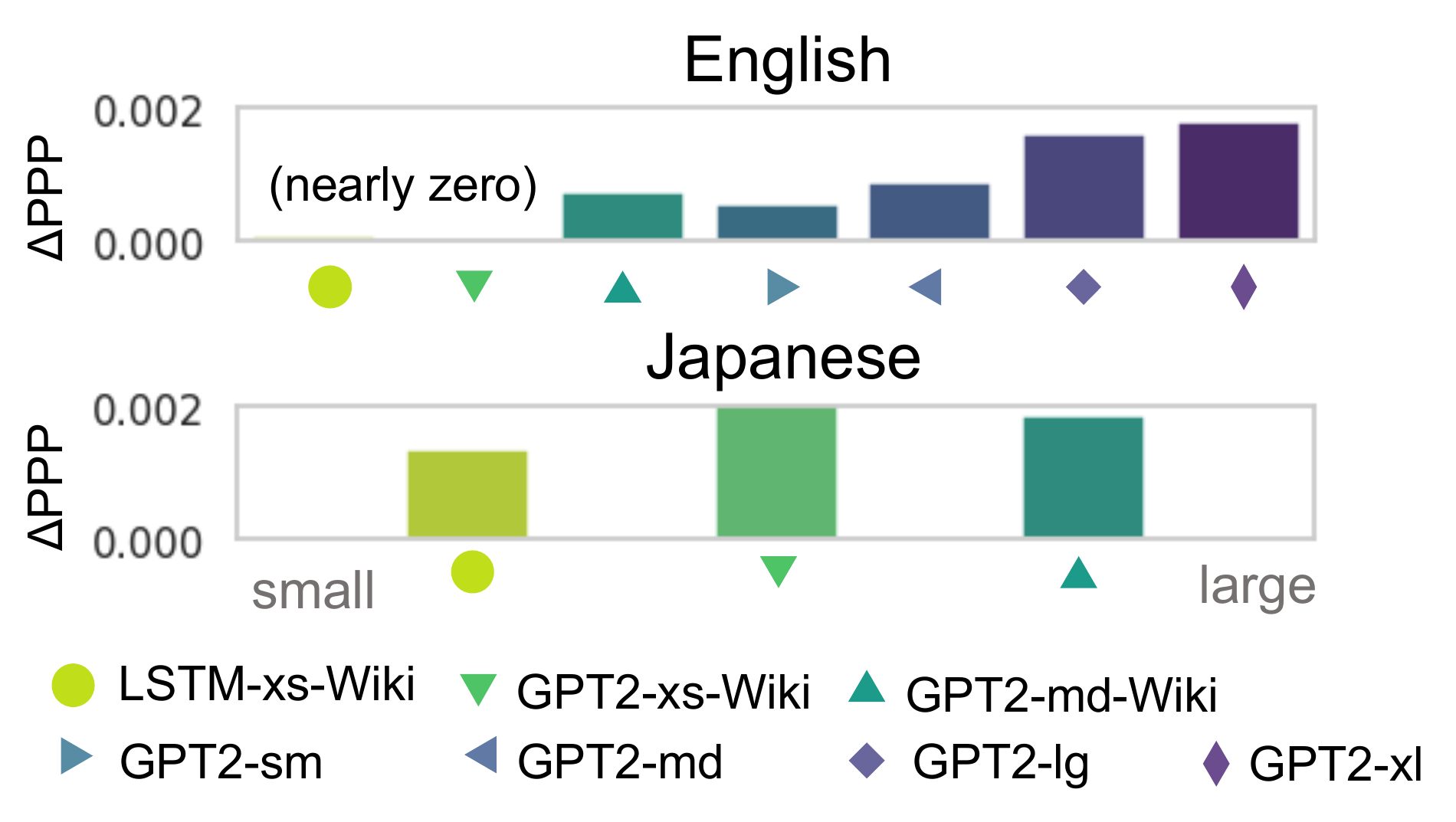}
    \hspace{-1.5cm}
　  \caption{Increase in PPP (from the full-gram to 2-gram settings) in each model type (ordered by their parameter size). The bar colors correspond to those in Figure~\ref{fig:ppp}.}
    \label{fig:delta_ppp}
          \setlength{\belowcaptionskip}{-30pt}
\end{figure}

\begin{figure*}[t]
    \centering
      \includegraphics[width=\linewidth]{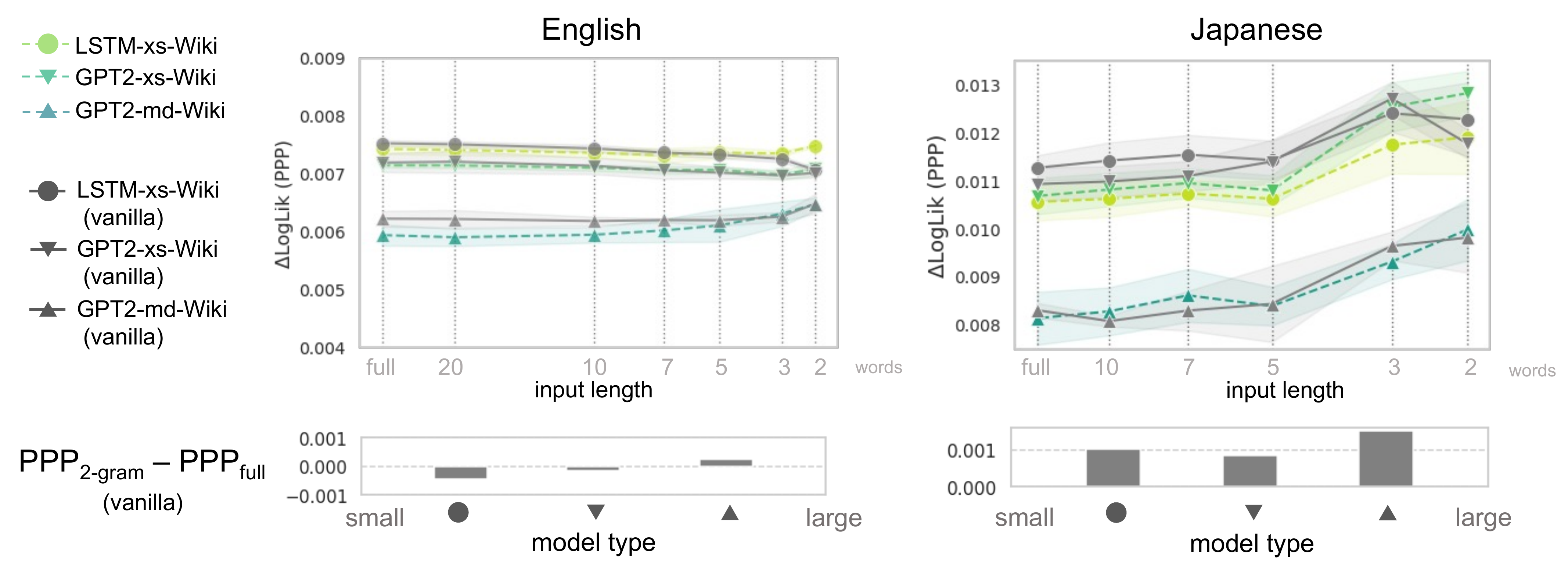}
      \caption{Reproduction of Figures~\ref{fig:ppp} and~\ref{fig:delta_ppp} using LMs without training setting modifications (Section~\ref{subsec:main_results}).
      The results from Wiki-LMs with the modification (colored) and without the modification (gray) are overlayed.
      In the line charts, the X-axis indicates input length, and the Y-axis indicates PPP.
      The bottom bar charts show the increase in PPP (from full-gram to 2-gram setting) of the modified LMs.
    }
     \label{fig:training_gap}
    \setlength{\belowcaptionskip}{-30pt}
\end{figure*}

\subsection{PPP and input length}
\label{subsec:main_results}

\paragraph{Shorter context improved or did not decrease human likeness.}
The PPP of $n$-gram surprisal in relation to input length $n$ is shown in Figure~\ref{fig:ppp} and Table~\ref{tbl:n-gram_lossy}.
The English results show that using a shorter context improved (OpenAI GPT-2s) or did not hurt (Wiki-LMs) their human likeness.
Notably, we also conducted experiments using probabilistic versions of the noise in Appendix~\ref{app:probabilistic_noise}, yielding consistent results.

Note that the tininess of the values in Figure 1 and Table 4 (e.g., 0.0074 v.s. 0.0056 in GPT2-xl) does not imply that the difference is valueless, but this is just because the score is divided by the number of data points (e.g., 212,649 in the Dundee corpus) to facilitate inter-corpora comparison.
As a statistical test, we compared the by-token squared residual errors from 2-gram models with those from full-context models using paired permutation tests (p=$0.05$).
The short context, 2-gram models had significantly smaller fitting errors than the full context models ($p<0.001$) in using relatively large LMs (GPT2-md-Wiki, GPT2-sm, GPT2-md, GPT2-lg, and GPT2-xl); smaller LMs (LSTM-xs-Wiki, and GPT2-xs-Wiki) have no significant differences ($p\sim 0.4$).

Notably, we also observed that \textbf{larger} GPT-2s have \textbf{less} human-like behavior in the $\mathrm{full}$ setting (right-most column in Table~\ref{tbl:n-gram_lossy}).
This trend was weakened by introducing our context limitation.

\paragraph{{Cross-linguistic consistency.}}
Figure~\ref{fig:ppp} and Table~\ref{tbl:n-gram_lossy} also show the cross-linguistic generality of the short context advantage. 
The short context was more clearly favored in Japanese than in English.
Using the same method as the English experiments, we performed the significance tests; 2-gram models exhibited smaller fitting errors ($p\sim0.001$) in all the Japanese LM settings.
The language-dependent differences are further investigated in Section~\ref{sec:exp3}.

\paragraph{The larger the LM, the greater the increase in PPP when limiting context access.}
Figure~\ref{fig:delta_ppp} shows the PPP increase in each LM class by context limitation (PPP at 2-gram minus PPP at full-gram).
The bars were ordered by the model parameter size (small $\xrightarrow{}$ large).
We found a clear trend that larger LMs become human-like by a larger margin because of context limitation; larger full-context LMs deviate more from human-like context access.

We statistically tested whether the gain by context limitation (full-context v.s. bigram) was larger in the largest LMs (GPT2-md in Japanese and GPT2-xl in English) than in the smallest LMs (LSTM-xs). 
Specifically, we compared the by-token decrease in squared residual errors; the large model exhibited a larger error decrease than the small model ($p=0.024<0.05$ in Japanese, and $p<0.001$ in English). 
In addition, the rank correlation between model size and PPP gain by context limitation was 0.50 in Japanese and 0.96 in English.

\paragraph{General effectiveness of surprisal.}
Note that, in all the LMs, the PPP scores (equivalent to $\Delta$logLik) were significantly higher than 0 with the chi-square test ($p<10^{-31}$ even in the worst case); surprisal was an effective factor as existing studies reported.
On top of this, we newly showed that their effect size differs due to the context limitation levels.

\subsection{Does the potential training-inference mismatch bias our results?}
\label{subsec:train_settings}

\textbf{Vanilla LMs slightly underestimate the short-context advantage.}
We additionally trained Wiki-LMs (LSTM-xs-Wiki, GPT2-xs-Wiki, and GPT2-sm-Wiki) without the data modification handling the training-inference gap (Section~\ref{subsec:wiki-lms}) (henceforth; vanilla LMs).
Figure~\ref{fig:training_gap} shows the results of the models with and without the training modification.
The vanilla LMs slightly underestimated the short-context advantage; the PPP of 2-gram surprisal improved when we adopted the modified training.
That is, mitigating the train-inference gap made clearer the trend that \textit{context limitation increases PPP}.
Carefully training n-gram neural LMs could be a way to create more human-like computational model.

\section{Analyses}
\label{sec:exp3}

In our experiments, we merely deleted distant contexts regardless of linguistic factors.
However, this design is somewhat counter-intuitive in the sense that humans are assumed to completely forget the distant context during reading.
To gain insights into a more sophisticated noise design, we observed \textit{in which constructions longer/shorter contexts improved simulation of human gaze duration.}

\subsection{Settings}
\paragraph{Quantifying long context effect.}
To quantify the long context advantage for each data point, we compared the squared residual (fitting) errors of the regression models we used to compute PPP in Section~\ref{sec:exp1}.
Note that the larger the squared residual error is, the worse the model fit with the target variable (gaze duration).

Specifically, we contrasted the two regression models with different context access: (i) the model with 2-gram surprisal, and (ii) the model with $\mathrm{full}$ context surprisal.
For each data point $d$, we measured the effectiveness of long context (ELC) in explaining gaze duration.
Specifically, the difference between the squared residual errors by the regression models with 2-gram surprisal $r_\mathrm{2}(d)$ and $\mathrm{full}$ surprisal $r_\mathrm{full}(d)$ was computed:
\begin{align}
    \mathrm{ELC}(d) = r_\mathrm{2}(d) - r_\mathrm{full}(d) \:\:.
    \label{eq:correl_cl}
\end{align}
\noindent
Here, a \textbf{high} ELC value indicates that reading times on $d$ were \textbf{better} simulated with \textbf{long} context ($r_\mathrm{full}(d)\downarrow$); \textbf{worse} simulated with \textbf{short} context ($r_\mathrm{2}(d)\uparrow$).
The aim of this section is to find the data points with a high ELC value.
In the following analyses, we used all the models from Section~\ref{subsec:main_results}, and ELC scores for each data point were averaged across all the LMs.

\paragraph{Dependency structure.}
Human context access has typically been discussed with respect to syntactic structure~\cite{Gibson1998LinguisticDependencies,Demberg2008DataComplexity}; we first explored the interactions between context limitation advantage and syntactic dependencies.
We analyzed two syntactic factors: (i) \textbf{dependency locality} and (ii) \textbf{dependency type}, where the dependency locality of a token denotes how far its syntactically related preceding items (i.e., with a direct dependency) are placed on average.
An example is as follows:

  \begin{dependency}[theme = simple, arc angle=30]
      \begin{deptext}[column sep=0.3em]
      (3) \& The \& boy \& over \& there \& had \& a \& cap.  \\
    \end{deptext}

    \depedge[-,label style={font=\large}]{6}{3}{3, \texttt{nsubj}}
    \depedge[-, gray]{2}{3}{}
    \depedge[-, gray]{3}{5}{}
    \depedge[-, gray]{5}{4}{}
    \depedge[-, gray]{6}{8}{}
    \depedge[-, gray]{8}{7}{}
    
  \end{dependency}

\noindent
Here, the dependency locality of ``had'' is three; note that the dependency direction was disregarded.

In the following analyses, we only used data points with potential long context access, i.e., those in the latter part of a sentence.\footnote{13th/7th or later words of a sentence in the DC/BE data (20,554 words/4,051 words) were used, based on the median of the word position in sentences (12 and 6 in the DC and BE corpus, respectively).}
After this filtering, the average dependency locality score was 2.5 and 2.6 in the DC and BE, respectively. 
Manual linguistic annotations were used in our analyses~\cite{Barrett2015-nd,Omura2018-xx}.

\subsection{Results}
\begin{figure}[t]
    \begin{minipage}[t]{\hsize}
    \centering
      \includegraphics[width=\linewidth]{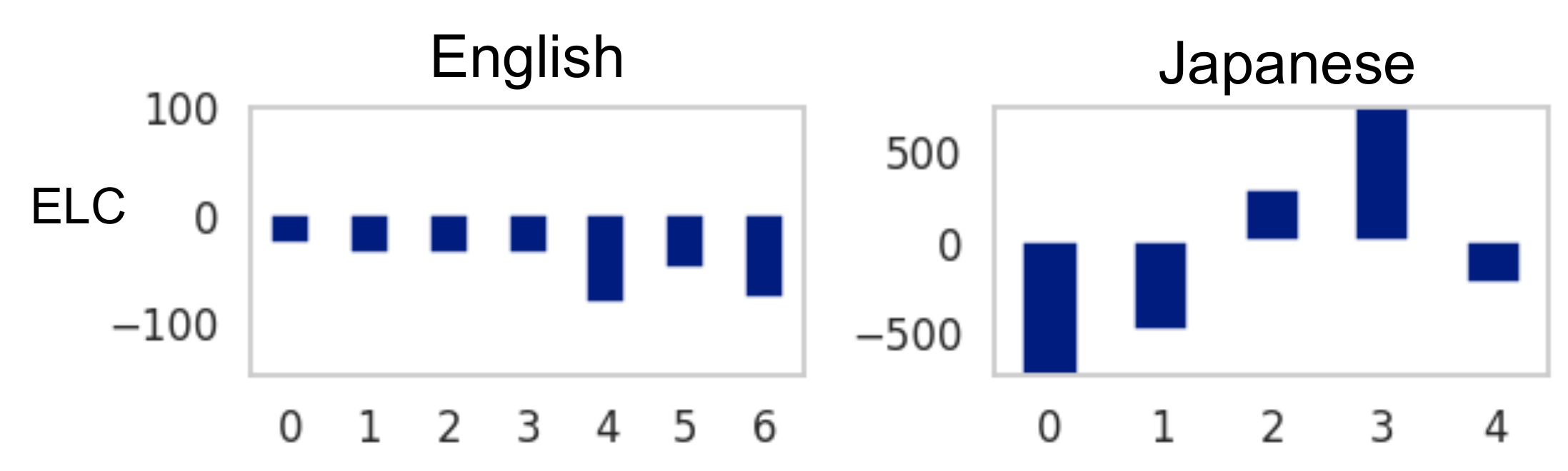}
      \vspace{-0.5cm}
      \subcaption{Relationship between dependency locality and the ELC scores. The X-axis corresponds to  dependency locality (e.g., the group ``3'' denotes the data points with the locality score of three). The Y-axis denotes the ELC score for each group.
    }
    \label{fig_dep_length}
    \vspace{0.5cm}
    \end{minipage} \\
        \begin{minipage}[t]{\hsize}
    \centering
      \includegraphics[width=\linewidth]{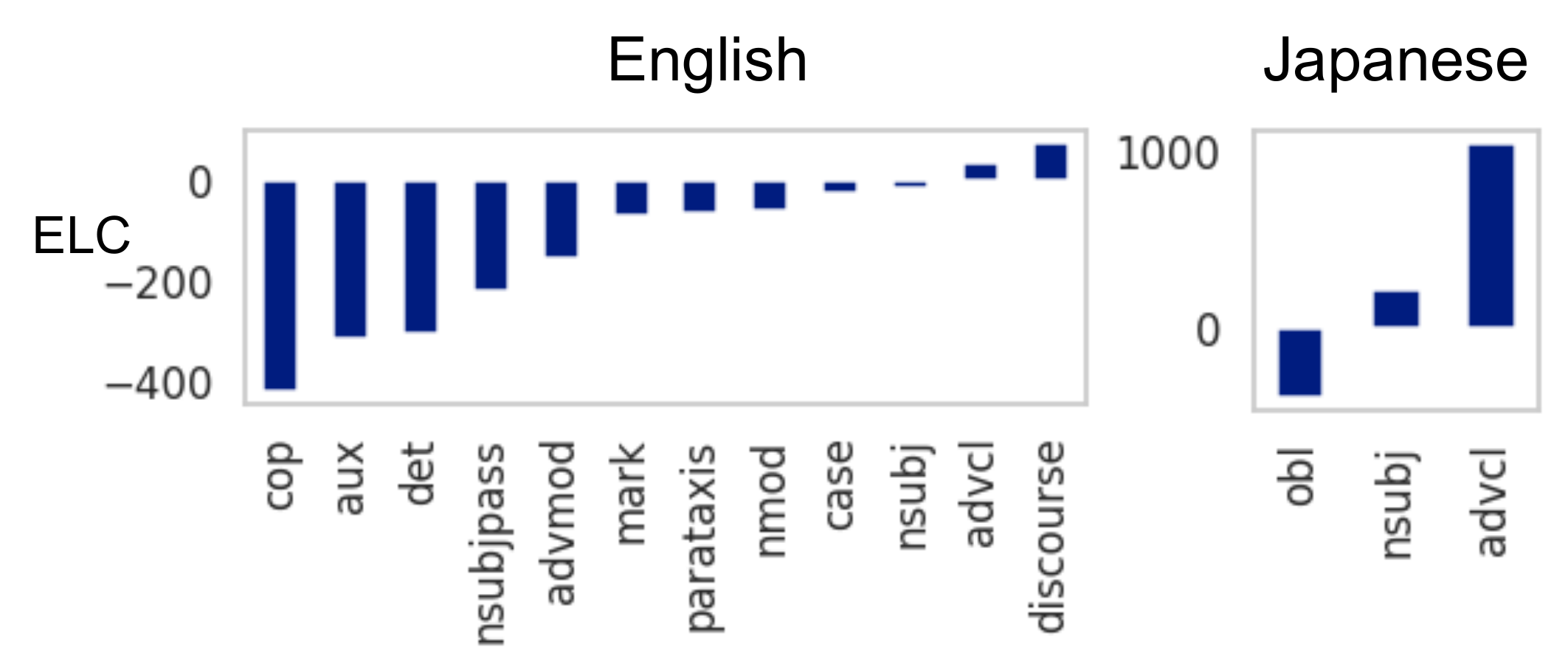}
      \subcaption{Relationship between dependency type and the ELC scores. The X-axis corresponds to  the dependency type. The Y-axis denotes the average ELC score for each group. Dependency types for which there are more than 100 long dependencies (locality$>$4) were included.
    }
        \label{fig_dep_type}
    \end{minipage} \\
    \caption{Relationships between syntactic factors and the ELC scores.
    }
      \label{fig:locality}
\end{figure}

\paragraph{Dependency locality.}
We first grouped the data points by their dependency locality and calculated the average ELC scores for each group.
Figure~\ref{fig_dep_length} shows the results.
Surprisingly, in the English data, there is no advantage in considering the long context for tokens with long dependencies.
By contrast, in the Japanese data, long context access contributed to simulating reading time for tokens with a moderate (two or three) dependency length, but not for long dependency locality.
These imply that \textbf{the solution is more complex than simply using long context for words with long dependency.}

\paragraph{Dependency type.}
Does long context matter in specific syntactic constructions?
We categorized the data points by their dependency type to their preceding syntactically related items and calculated the averaged ELC score for each group.\footnote{Here, we only focus on the dependencies with more than four distances to explore potential long context access.}

Figure~\ref{fig_dep_type} shows that different dependency types are associated with different ELC scores.
For example, the discourse type in English have relatively larger ELC scores; long context input is necessary to simulate its gaze duration.
Figure~\ref{fig_dep_type} also suggests that such context-favoring (with high ELC) dependency types are different between English and Japanese.
These findings imply that \textbf{the LM-human context access gap occurred in specific syntactic constructions in each language.}

One-way ANOVA revealed that the average ELC scores for each dependency type significantly varied ($p=0.029<0.05$ in English, $p=0.038<0.05$ in Japanese), suggesting that the variation of the ELC score is related to certain constructions.
More specifically, we compared the ELC distribution between the categories with the highest and lowest average ELC scores (discourse vs. cop in English, and advcl vs. obl) using an unpaired t-test.
The test exhibited a significant difference ($p=0.012<0.05$ in English, and $p=0.019<0.05$ in Japanese).
Note that if the test is repeated for other dependency-length/type pairs, multiple comparison problems should occur; some counteractions, such as Bonferroni correction should be applied, and a more conservative conclusion can be reached.

\section{Discussion}
\label{sec:limitations}

\subsection{Interpretations of the main results}

We observed that simply deleting distant context improved LMs' PPP---\textit{as context decreased, LMs became more human-like}.
We finally discuss several potential interpretations of our results.

One interpretation is that our results supported \textbf{the dominance of short context access in human sentence processing}.
In this sense, our findings emphasized that explicitly incorporating principles from the memory-based account of human sentence processing is still necessary for simulating human sentence processing despite the success of modern LMs in cognitive modeling~\cite{Wilcox2020OnBehavior,schrimpf2020neural}.
Notably, there are several other theories on human working memory; sparse allocation of elements incurring memory load~\cite{Gibson2000TheComplexity}, hierarchical memory operations~\cite{Van_Schijndel2013-sn}, and cue-based memory  retrieval~\cite{Lewis2005-hp}.
Incorporating these perspectives into context-limited LMs could be an interesting future direction.

Another possibility is that \textbf{identifying the cause of the LM-human gap as context limitations is over-claiming}; our study alone did not rule out some potentially confounding factors.
For example, increasing the softmax temperature when LMs compute the next-word distribution may induce a similar effect to our context limitation with respect to that both modifications make LMs less confident about the upcoming word (if temperature matters, the linear relationship between surprisal and cognitive load may be doubted first).
Further exploring such factors will be an important investigation.

There is also a possibility that \textbf{the eye movement data only reflected local, shallow aspects in human sentence processing}.
Similarly, \citet{Gauthier2019-nj} obtained somewhat counter-intuitive results implying that word order is not important information in sentence processing---bag-of-words (i.e., not word-order-aware) models fit fMRI data surprisingly well.
They concluded that their results may stem from shortcomings of the measurement method along with the possibility of humans' good-enough processing.
Exploring the advantage of context limitation in various types of reading behavior data and/or using other text materials (e.g., including more complex constructions) is also a line of future research.

\paragraph{Is the 2-gram advantage counter-intuitive?}
If the dominance of short context access in human reading is accepted once, some readers might be confused that the 2-gram context access sounds too severe.
Again, as a strong word frequency effect does not deny context-dependent processing, our results also did not decline long context access and did not claim that the human language processing model is 2-gram LMs.
Exploring the \textit{interactions} of short- and long-context effects should be an interesting investigation.

Nevertheless, such severe memory limitations during reading might be consistent with the memory-based explanation for the linguistic universals in sentence structures such as the preferences toward consistent head directions, specific word order (e.g., short-before-long order), and projective structures~\cite{Futrell2020DependencyOrder}.
Such phenomena are typically explained by the humans' preferences toward short dependencies; here, those are sometimes a matter of severe choices, such as the preference for an average dependency length of 1 over 2 (Intuitively, Example (4) is preferred over Examples (5) and (6)):

\begin{dependency}[theme = simple]
  \begin{deptext}[column sep=0.3em]
  (4) \& A \&  B  \&  C  \& D  \& E \& avg. dep len.=1.25 \\
\end{deptext}

\depedge[label style={font=\large}]{2}{4}{2}
\depedge[label style={font=\large}]{3}{4}{1}
\depedge[label style={font=\large}]{4}{5}{1}
\depedge[label style={font=\large}]{5}{6}{1}

\end{dependency}

\begin{dependency}[theme = simple]
  \begin{deptext}[column sep=0.3em]
  (5) \& A \& B \& C \& D \& E \& avg. dep len.=2.25 \\
\end{deptext}

\depedge[label style={font=\large},left=8pt]{2}{6}{4}
\depedge[label style={font=\large},left=3pt]{6}{3}{3}
\depedge[label style={font=\large},right=5pt]{3}{4}{1}
\depedge[label style={font=\large},left=2pt]{4}{5}{1}

\end{dependency}

\begin{dependency}[theme = simple]
  \begin{deptext}[column sep=0.3em]
  (6) \& A \& B \& C \& D \& E \& avg. dep len.=1.50 \\
\end{deptext}

\depedge[label style={font=\large}]{2}{4}{2}
\depedge[label style={font=\large}]{3}{5}{2}
\depedge[label style={font=\large}]{4}{5}{1}
\depedge[label style={font=\large}]{5}{6}{1}

\end{dependency}

If one reasons these principles to the constraints of humans' cognitive resources, perhaps it makes sense that humans conduct syntactic processing with such a severe working memory that the immediately preceding word/phrase highly explains the cognitive load to the upcoming word.

 \subsection{PPP and next-word prediction accuracy}
\label{subsubsec:perplexity}

Lastly, we discuss the connection to reports on cognitive modeling with LMs---better next-word prediction ability of LMs indicates their better PPP~\cite{fossum-levy-2012-sequential,Goodkind2018PredictiveQuality,Wilcox2020OnBehavior}.\footnote{There are also some counter-arguments~\cite{Hao2020-wo,Oh2021-ln,kuribayashi-etal-2021-lower,Anonymous2022-ui}.}
Our results in Section~\ref{subsec:main_results} might be conflicting with them; LMs with relatively \textbf{worse} prediction accuracy (less context access) exhibit \textbf{better} PPP.
See Appendix~\ref{app:ppl_ngram} for next-word prediction accuracy of LMs.

\begin{figure}[t]
    \centering
      \includegraphics[width=\linewidth]{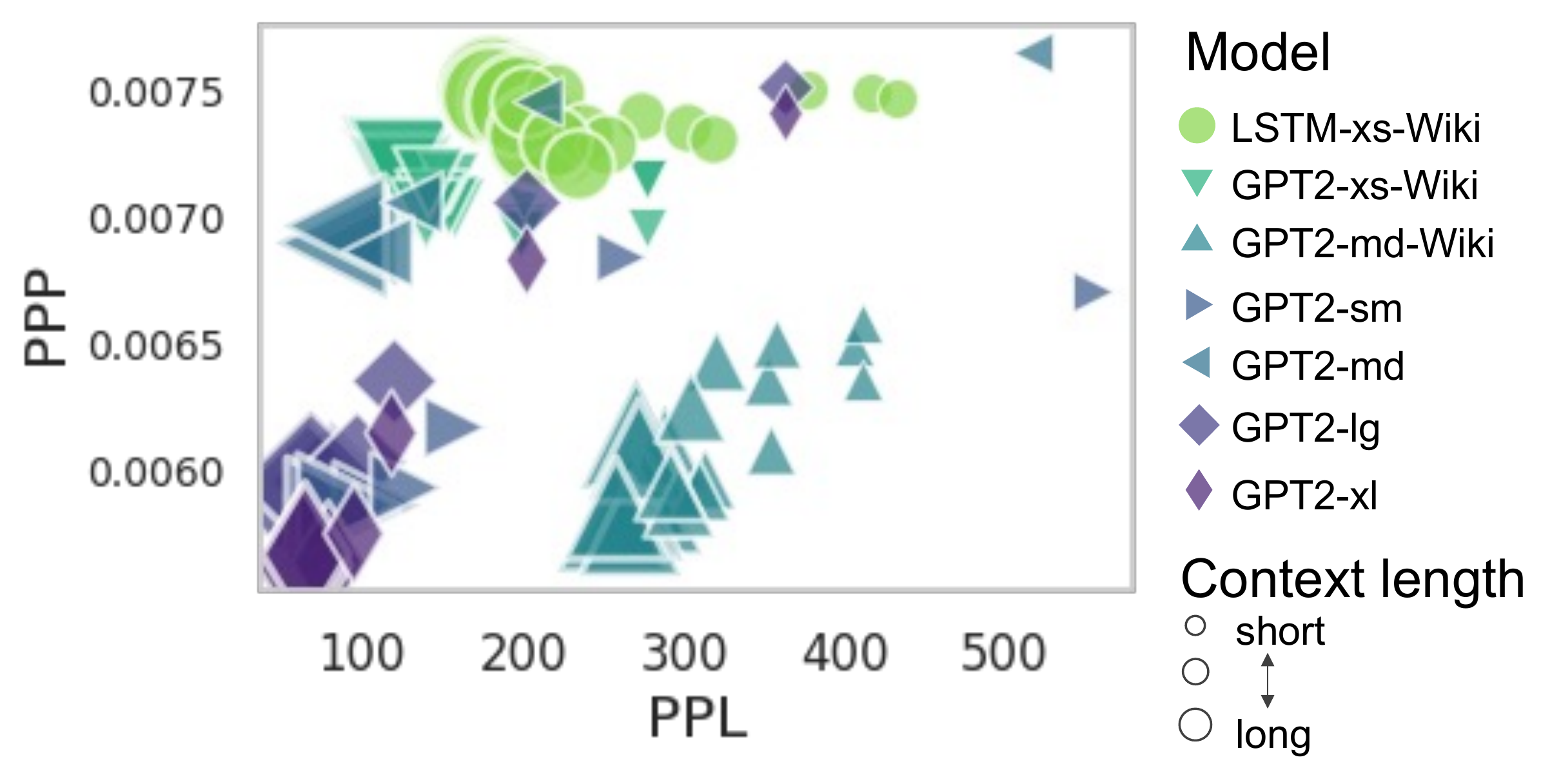}
      \caption{Relationship between PPP and perplexity (PPL) drawn using the English LMs targeted in Section~\ref{subsec:main_results}.
      Each point corresponds to each configuration of the n-gram surprisal computation; marker color and shape present the LM architectures, and larger markers correspond to longer context access.
      }
      \setlength{\belowcaptionskip}{-30pt}
      \label{fig:ppl}
\end{figure}

\paragraph{Results.}
In fact, there is no clear relationship between PPP and next-word prediction accuracy (perplexity; PPL) of the LMs used in Section~\ref{subsec:main_results} (Figure~\ref{fig:ppl}; Appendix~\ref{app:ppl_ppp_ja} exhibits Japanese results).
The results show that LMs with nearly the same next-word prediction accuracy could show different PPP values.
Furthermore, Pearson's correlation between PPP and PPL was even positive ($r=0.15$).
These observations corroborated the conclusion that the \textbf{PPL alone is not a good indicator of PPP}; different means of controlling PPL (e.g., context length vs. other factors existing studies focused on) could show different PPP-PPL relationships.

\section{Conclusions}
There has been little investigation of the cognitive plausibility of context-limited modern LMs.
Our experiments using the input-controlled neural LMs have shown that short context LMs simulate human reading behavior surprisingly well, emphasizing the LM-human gap in context access.
Further analysis has shown that the gap could be associated with specific syntactic constructions; injecting syntactic bias into LMs' context access could be one way to make LMs more human-like.
This study has also asserted that the use of a modern LM popular in NLP as-is is not always a natural choice in cognitive modeling.

\section*{Limitations}
As discussed in Section~\ref{sec:limitations}, this study alone could not comprehensively explain the cause of the LM-human discrepancies.
Nevertheless, our observation itself could advance the step toward understanding the relationship between human sentence processing and computational models typically developed in NLP, which is a central theme in the long history of artificial intelligence and the cognitive science of language.

This study was scientifically motivated to understand humans and language; this could sound like less impact on engineering-oriented efforts (e.g., solving real-world problems accurately).
However, simulating human cognitive load during reading is directly associated with automatic text readability assessment.
In addition, our study implies that human sentence processing could be performed with more efficient context access than modern LMs.
This encourages the development of language processing models with increased efficiency; this is related to the sustainability issues, such as the environmental impact of creating gigantic NLP models.

\section*{Ethical considerations}
This study explored the relationship between the LM-computed complexity measures and human reading behaviors.
Human subjects' privacy information in the eye-tracking data was anonymized.
We did not find any other ethical concerns; as a somewhat minor point, the LMs used in our experiments might be biased by the data we used (i.e., Wikipedia and Web data), although these follow the commonly used settings in the NLP research.

\section*{Acknowledgements}
We would like to thank Jun Suzuki and Ryo Yoshida for their valuable advice and appreciate the members of the Tohoku NLP Group for their constructive comments.
This work was supported by Grant-in-Aid for JSPS Fellows Grant Number JP20J22697,  JST PRESTO Grant Number JPMJPR21C2, and JST CREST Grant Number JP-MJCR20D2, Japan.

\newpage

\bibliography{Mendeley}
\bibliographystyle{acl_natbib}

\clearpage
\appendix
\section{Psychometric predictive power and regression models}
\label{app:feature}

Psychometric predictive power refers to the similarity between (lossy-context) surprisal and human gaze duration, calculated using a linear mixed-effects regression~\cite{JSSv067i01}.
First, gaze duration (\texttt{GD}) is modeled by the following formula: 

\vspace{-0.3cm}
{\small
\begin{align}
 \begin{split}
    \texttt{GD} &\sim \texttt{surprisal} + \texttt{surprisal\_prev\_1} \\
    &+ \texttt{surprisal\_prev\_2} + \texttt{freq}*\texttt{length} \\
    &+ \texttt{freq\_prev\_1}*\texttt{length\_prev\_1} \\
    &+ \texttt{screenN} + \texttt{lineN} + \texttt{segmentN} \\
    &+ \texttt{(1|article)} + \texttt{(1|subj)} \:\:.
    \label{eq:regression}
 \end{split}
\end{align}
}
\vspace{-0.3cm}

\noindent
Table~\ref{tbl:feature_name} shows the descriptions for the factors used in the above formulation. 
Then, a baseline regression model without the \texttt{surprisal}, \texttt{surprisal\_prev\_1}, and \texttt{surprisal\_prev\_2} terms from Eq.~\ref{eq:regression} is trained additionally.
We calculated the per-token average of the log-likelihood difference ($\Delta$LogLik) between the two regression models.

\section{Probabilistic erasure noise}
\label{app:probabilistic_noise}

\citet{Futrell2017noisy} and \citet{Futrell2020Lossy-ContextProcessing} suggested that a linear probabilistic erasure noise (LPEN), where more distant items are more likely to disappear as opposed to a constant cutoff point with n-grams, might be a plausible design of input limitations.
We examined whether such a probabilistic nature of noise design substantially affects our conclusions.
Within our experimental settings, there is no substantial difference in the results regardless of the probabilistic nature of noise.

\paragraph{Methods.}
To implement LPEN, we erased the $j$-th nearest word in the context with a probability of $\mathrm{min}(j*a, 1)$, here $a>0$.
We initially observed that erasing too close context hindered human-like behavior; we also introduced an always-present portion of the context ($l$ nearest words) and applied noise only on farther words.
That is, the probabilistic erasure noise is only applied to [$w_0, \cdots, w_{i-l-1}$].
Assuming $a=0.25$, $w_{i-l-1}$ is then erased with a probability of $0.25$, $w_{i-l-2}$ is erased with a probability of $0.5$, and so on, while the $l$ nearest words to the target are left intact.
We compared the PPP of surprisal with $l \in \{2, 3, 5, 7, 10, 20\}$ and $a \in \{0.5, 0.25, 0.125, 0.0625\}$.

\begin{figure*}[t]
    \centering
      \includegraphics[width=\linewidth]{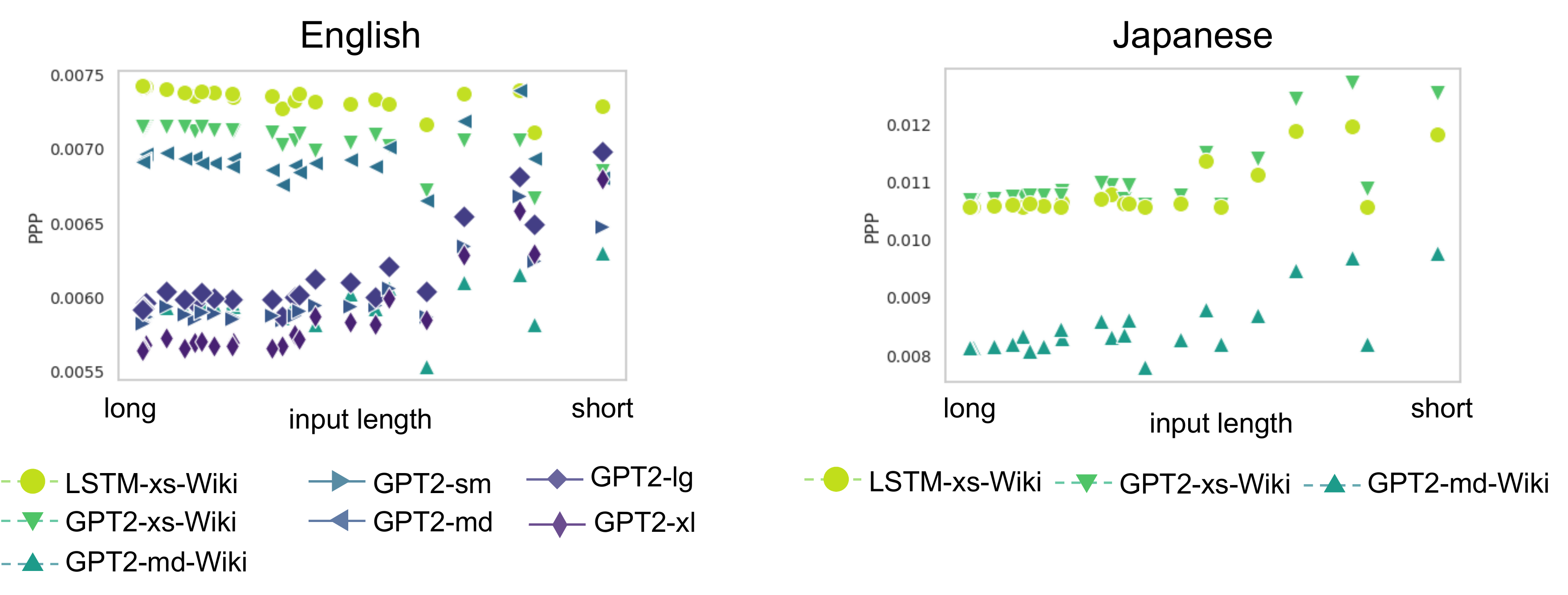}
      \caption{Relationship between the strength of input noise (X-axis) and PPP (Y-axis) under the probabilistic erasure noise settings.
        The results with various noise settings ( $L\times A$ where $L=\{2, 3, 5, 7, 10, 20\}$ and $A=\{0.5, 0.25, 0.125, 0.0625\}$) are summarized with respect to average input length in each setting.
        The marker color and shape correspond to LM architectures.
      }
     \label{fig:lossy_ppp}
\end{figure*}

\paragraph{Context limitation did not change or improved PPP.}
The results are shown in Figure~\ref{fig:lossy_ppp}.
The trends were similar to those using discrete context noise (Figure~\ref{fig:ppp}): (i) context limitation did not change or improved PPP and (ii) larger LMs have larger PPP gain due to context limitation.

\begingroup
\begin{table}[t]
\centering
{\footnotesize
\begin{tabular}{p{1.5cm}lp{3.5cm}} \toprule
Factor & Type & Description \\ 
\cmidrule(lr){1-1} \cmidrule(lr){2-2} \cmidrule(lr){3-3}
\texttt{surprisal} & num & (lossy-context) surprisal calculated by LMs \\
\texttt{GD} & num & reading time (first pass gaze duration)\\
\texttt{article} & factor & article ID\\
\texttt{screenN} & int & screen display order \\
\texttt{lineN}& int & the serial number of line the segment is displayed \\
\texttt{segmentN} & int & the serial number of segment in a screen \\
\texttt{sentN} & int & the serial number of sentence the segment belongs to\\
\texttt{tokenN} & int & the position of segment in sentence \\
\texttt{length} & int & number of characters \\
\texttt{freq} & num & geometric mean of the frequencies of subword constituents in a segment \\
\texttt{subj} & factor & participant ID \\
\bottomrule
\end{tabular}
}

    \caption{Factor names and their description.}
    \label{tbl:feature_name}
\end{table}
\endgroup

\section{Exclusion criteria for eye movement data}
\label{app:data}

We excluded outliers following~\citet{Goodkind2018PredictiveQuality}.
Specifically, we excluded the data points meeting any of the following criteria in the English experiments, and those meeting (a), (c), or (e) in the Japanese experiments:
\begin{description}
\setlength{\parskip}{0cm} 
\setlength{\itemsep}{0.1cm}
    \item[(a)] has zero gaze duration or beyond three standard deviations
    \item[(b)] contains punctuation
    \item[(c)] contains numeric characters
    \item[(d)] the next segment has punctuation or numeric characters 
    \item[(e)] is the first segment in a line
    \item[(f)] is the last segment in a line
\end{description}

\noindent
We included data points meeting (b) and (f) in the Japanese data out of concern that excluding them disregards the data points for the main verb, regarding the verb-final Japanese construction (punctuation is included in a \textit{bunsetsu}).
Note that in the Japanese data, the first/end word in a line correspond to first/end word in a sentence (sentences are presented line by line.).
Similarly, (d) substantially reduces the Japanese data points, and the inter-segment-level influence of special symbols would be less likely than in English considering that \textit{bunsetsu} is a relatively larger unit than the English word.

\section{Hyperparameters of LMs}
\label{app:hyper}

\begin{table*}[ht]
    \centering
\begin{minipage}[t]{\hsize}
\renewcommand{\arraystretch}{0.5}
    \centering
    \begin{tabular}{p{3cm}lp{4.5cm}} \toprule
     \multirow{9}{1cm}{Fairseq model} & architecture & transformer\_lm \_gpt2\_small \\
      & adaptive softmax cut off & 50,000, 140,000 \\
      & share-decoder-input-output-embed & True \\
      & embed\_dim & 1,024 \\
      & ffn\_embed\_dim & 4,096 \\
      & layers & 24 \\
      & heads & 16 \\
      & dropout & 0.1 \\
      & attention\_dropout & 0.1 \\
    \cmidrule(lr){1-1} \cmidrule(lr){2-2} \cmidrule(lr){3-3}
    \multirow{5}{*}{Optimizer} & algorithm & AdamW \\
    & learning rates & 5e-4 \\
    & betas & (0.9, 0.98) \\
    & weight decay & 0.01 \\
    & clip norm & 0.0 \\
    \cmidrule(lr){1-1} \cmidrule(lr){2-2} \cmidrule(lr){3-3}
    \multirow{3}{3cm}{Learning rate scheduler} & type & inverse\_sqrt \\
    & warmup updates & 4,000 \\
    & warmup init lrarning rate & 1e-7 \\
    \cmidrule(lr){1-1} \cmidrule(lr){2-2} \cmidrule(lr){3-3}
    \multirow{2}{*}{Training} & batch size & 61,440 tokens \\
    & sample-break-mode & none \\ \bottomrule
        \end{tabular}
        \subcaption{GPT2-md-Wiki.}
        \label{tbl:hyperparam_tl}
        \vspace{0.2cm}
\end{minipage}

\begin{minipage}[t]{\hsize}
\renewcommand{\arraystretch}{0.5}
    \centering
    \begin{tabular}{p{3cm}lp{4.5cm}} \toprule
     \multirow{9}{1cm}{Fairseq model} & architecture & transformer\_lm\_gpt \\
      & adaptive softmax cut off & 50,000, 140,000 \\
      & share-decoder-input-output-embed & True \\
      & embed\_dim & 384 \\
      & ffn\_embed\_dim & 2,048 \\
      & layers & 8 \\
      & heads & 6 \\
      & dropout & 0.1 \\
      & attention\_dropout & 0.1 \\
    \cmidrule(lr){1-1} \cmidrule(lr){2-2} \cmidrule(lr){3-3}
    \multirow{5}{1cm}{Optimizer} & algorithm & AdamW \\
    & learning rates & 5e-4 \\
    & betas & (0.9, 0.98) \\
    & weight decay & 0.01 \\
    & clip norm & 0.0 \\
    \cmidrule(lr){1-1} \cmidrule(lr){2-2} \cmidrule(lr){3-3}
    \multirow{3}{3cm}{Learning rate scheduler} & type & inverse\_sqrt \\
    & warmup updates & 4,000 \\
    & warmup init learning rate & 1e-7 \\
    \cmidrule(lr){1-1} \cmidrule(lr){2-2} \cmidrule(lr){3-3}
    \multirow{2}{1cm}{Training} & batch size & 61,440 tokens \\
    & sample-break-mode & none \\ \bottomrule
        \end{tabular}
        \subcaption{GPT2-xs-Wiki.}
        \label{tbl:hyperparam_ts}
        \vspace{0.2cm}
\end{minipage}

\begin{minipage}[t]{\hsize}
\renewcommand{\arraystretch}{0.5}
    \centering
    \begin{tabular}{p{3cm}lp{4.5cm}} \toprule
     \multirow{7}{1cm}{Fairseq model} & architecture & lstm\_lm \\
      & adaptive softmax cut off & 50,000, 140,000 \\
      & share-decoder-input-output-embed & True \\
      & embed\_dim & 400 \\
      & hiden\_size & 1,024 \\
      & layers & 2 \\
      & dropout & 0.1 \\
    \cmidrule(lr){1-1} \cmidrule(lr){2-2} \cmidrule(lr){3-3}
    \multirow{5}{1cm}{Optimizer} & algorithm & AdamW \\
    & learning rates & 1e-3 \\
    & betas & (0.9, 0.98) \\
    & weight decay & 0.01 \\
    & clip norm & 0.0 \\
    \cmidrule(lr){1-1} \cmidrule(lr){2-2} \cmidrule(lr){3-3}
    \multirow{3}{1cm}{Learning rate scheduler} & type & inverse\_sqrt \\
    & warmup updates & 4,000 \\
    & warmup init learning rate & 1e-7 \\
    \cmidrule(lr){1-1} \cmidrule(lr){2-2} \cmidrule(lr){3-3}
    \multirow{2}{3cm}{Training} & batch size & 20,480 tokens \\
    & sample-break-mode & none \\ \bottomrule
        \end{tabular}
        \subcaption{LSTM-xs-Wiki.}
        \label{tbl:hyperparam_lstm}
        \vspace{0.2cm}
\end{minipage}

\caption{Hyperparameters of the LMs.}
\label{tbl:hyper_params}
\end{table*}

Table~\ref{tbl:hyper_params} shows the hyperparameters of Wiki-LMs.
Training each LM took approximately three days on four GPUs (NVIDIA V100).
The hyperparameters of OpenAI GPT-2s could be shown in \url{https://huggingface.co/docs/transformers/model_doc/gpt2}.

As for the subword tokenization used in Wiki-LMs, we set character coverage to 0.9995.
Vocabulary size was set to 32,000 in English and 100,000 in Japanese, taking the rich characters and morphemes in Japanese into consideration.
Note that this difference results in Japanese Wiki-LMs having more parameters than English LMs. In the Japanese settings, LSTM-xs-Wiki has 54M (27M in English), GPT2-xs-Wiki has 55M (29M in English), and GPT2-md-Wiki has 404M (335M in English) parameters.

{\setlength\textfloatsep{0pt}
\begin{figure*}[t]
    \centering
      \includegraphics[width=\linewidth]{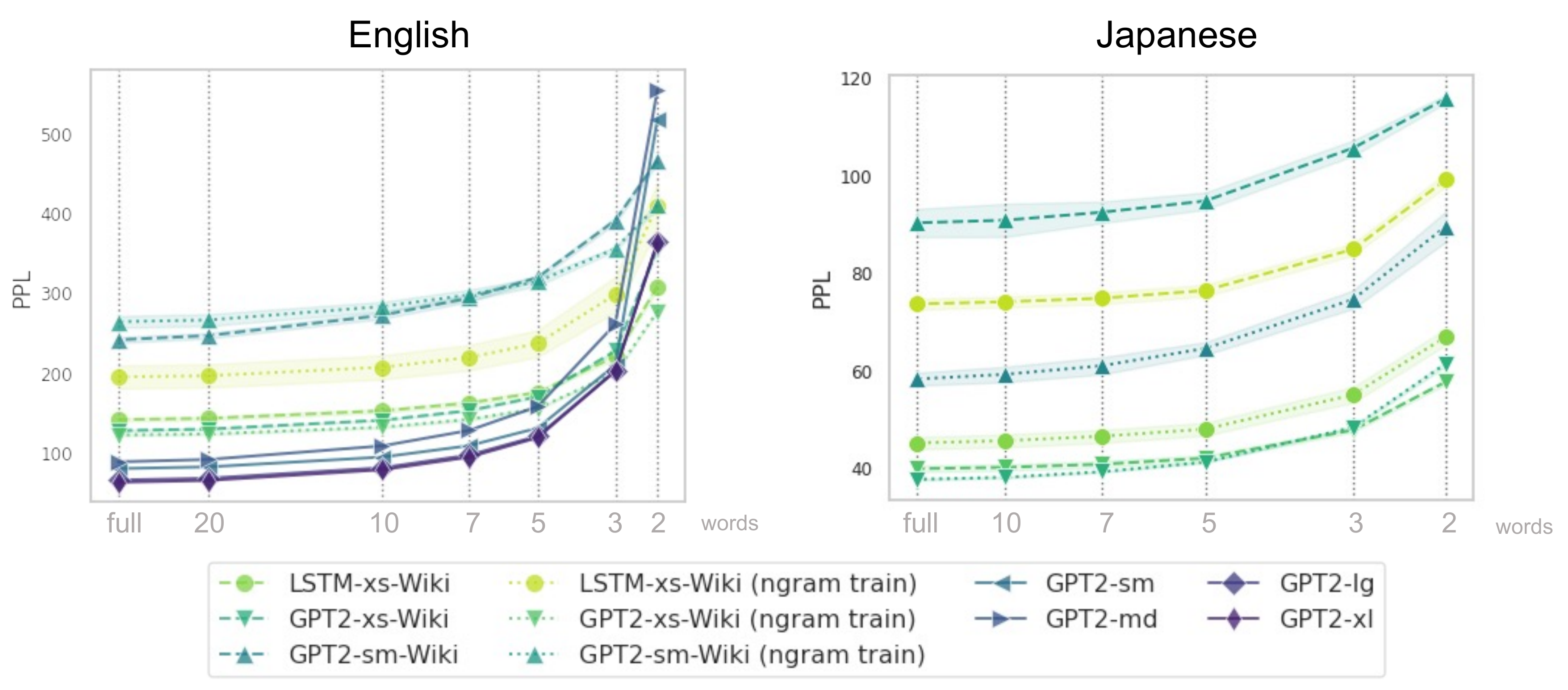}
      \caption{Relationship between the perplexity of n-gram LMs and input length. A monotonic relationship, the longer context LMs use, the lower perplexity LMs exhibit, is observed. 
      The colored areas show a 95\% confidence interval. The PPL was computed at the subword level; here, directly comparing the scale of Y-axis across languages is non-sense due to their different segmentation (e.g., vocabulary size).}
      \label{fig:ngram_ppl}
\end{figure*}
}

\section{Mitigating training-inference gaps}
\label{app:settings_modify}

As introduced in Section~\ref{subsec:wiki-lms}, we modified the training data to augment the data points, where LMs must predict the upcoming tokens from the middle of a sentence with severely limited context.
Specifically, we first split each ($i$-th) sentence, $s^i=[w_0^i, w_1^i, \cdots, w_n^i]$, in training data into two sub-sequences: [\texttt{<s>}, $w_0^i, \cdots, w_{k-1}^i$] and [\texttt{<b>}, $w_{k}^i, \cdots, w_n^i$].
Here, the breakpoint $k^i$ for the $i$-th sentence is sampled from the uniform distribution $U(0, |s^i|)$.
When $k=0$, the former sub-sequence is [\texttt{<s>}].
Then, the sub-sequences obtained from the whole corpus were randomly concatenated to create the modified training data (e.g., [$\cdots$, \texttt{<b>}, $w_{k^c+1}^c, \cdots, w_n^c$, \texttt{<b>}, $w_{k+1}^a, \cdots, w_n^a$, \texttt{<s>}, $w_0^l, \cdots, w_k^l$, $\cdots$]).
This modified data has two characteristics: (i) there is no dependency between before and after the special tokens (\texttt{<s>} and \texttt{<b>}), and (ii) to the \texttt{<b>} token, uniform prior about the token position within the sentence is set.
It is expected that training LMs with this data and computing the next word prediction (e.g., with the query of [\texttt{<b>}, $w$]) enables them to adequately compute the upcoming token (i) with severely limited context and (ii) without any token position prior for \texttt{<b>}.

Concretely, the lossy-context surprisal is computed by the modified LMs as follows:
\begin{align}
\nonumber
    &I_\mathrm{lossy}(w_i, c_{<i}, f) \\ 
    \label{eq:surprisal_modify}
    &= -\log p_{\theta}(w_{i}| \mathrm{\texttt{<b>}} \circ f([w_0, \cdots, w_{i-1}])) \:\:,
\end{align}

Only when the \texttt{<b>} token corresponds to the sentence initial position, \texttt{<s>} was set instead of \texttt{<b>}.

\section{Perplexity of n-gram LMs}
\label{app:ppl_ngram}

\paragraph{Perplexity (PPL):}

Perplexity (PPL), the inverse geometric mean of next-word probabilities $p(w_{i}|w_{<i})$ in a text that consists of $N$ symbols ($w_1, w_2, \cdots, w_N$), is a typical evaluation metric for the next-word prediction accuracy.
Given a LM $\theta$ and context noise design $f$, the perplexity under lossy-context settings is calculated as follows:

\begin{align}
    \text{PPL} &=\prod_{i=0}^N p_{\theta}(w_{i}| \mathrm{BOS} \circ f(c_{<i}))^{-\frac{1}{N}} \:\:.
    \label{eq:ppl}
\end{align}

\noindent
Low PPL indicates that the model and the context noise yielded accurate predictions about the upcoming signal.

\paragraph{Relationship between perplexity and context length:}
Figure~\ref{fig:ngram_ppl} shows the relationship between the perplexity of n-gram LMs and their average context length.
The PPL values are computed with the texts in the eye movement data.
A monotonic relationship, the longer context LMs use, the lower perplexity LMs exhibit, is observed.
This ensures that LMs with long context actually exploit the information in the added context to accurately predict the upcoming symbols.

\section{Next-word prediction accuracy and PPP in Japanese}
\label{app:ppl_ppp_ja}

\begin{figure}[t]
    \centering
      \includegraphics[width=\linewidth]{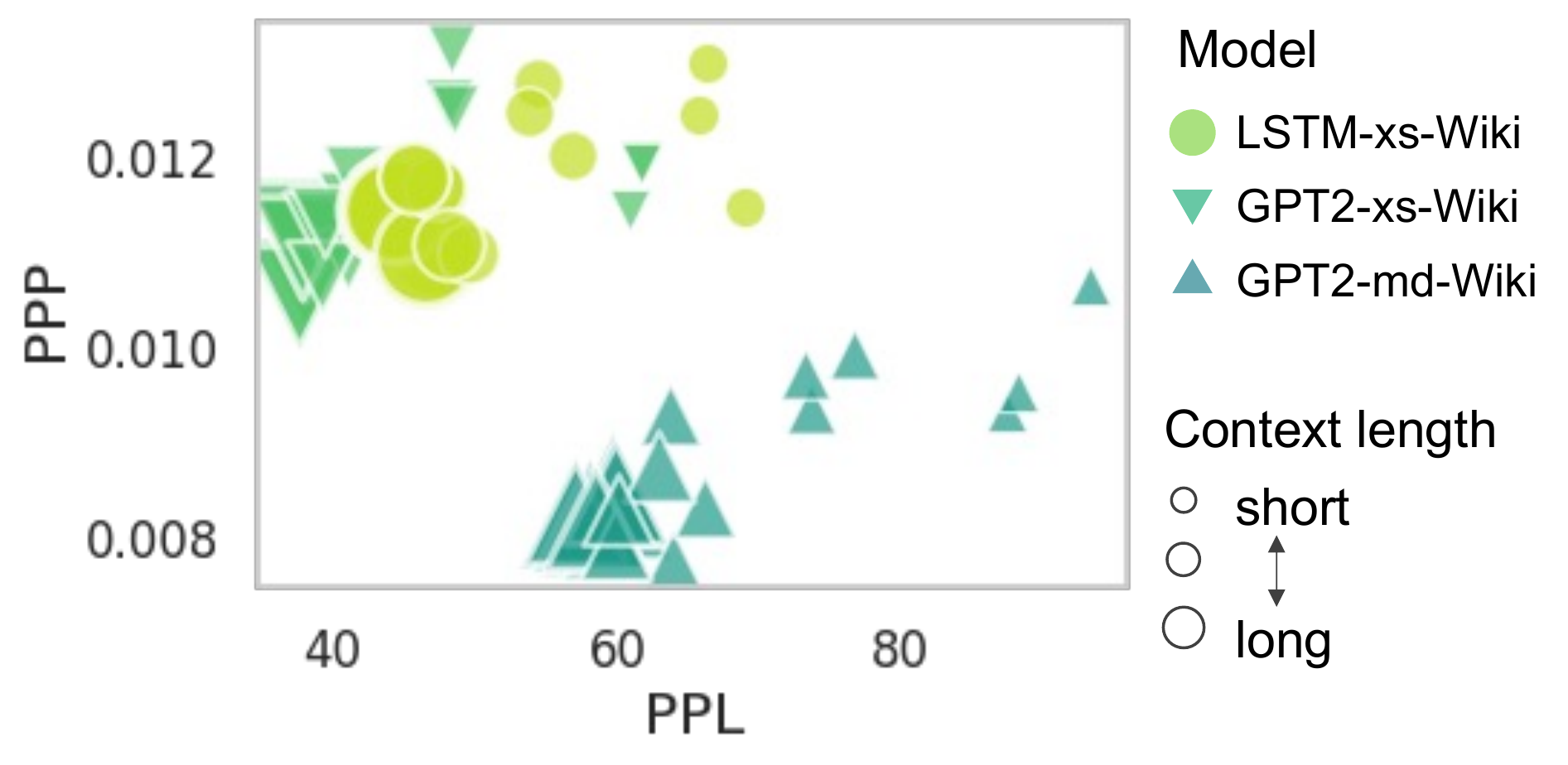}
      \caption{Relationship between PPP and perplexity (PPL) drawn using the Japanese LMs targeted in Section~\ref{subsec:main_results}.
      Each point corresponds to each configuration of the n-gram surprisal computation; marker color and shape present the LM architectures, and larger markers correspond to longer context access.
      }
      \setlength{\belowcaptionskip}{-30pt}
      \label{fig:ppl_ja}
\end{figure}

Figure~\ref{fig:ppl_ja} shows the relationship between PPL and PPP in the Japanese experiments.
Similar to the results of Section~\ref{subsec:main_results}, LMs with a similar PPL value exhibited different PPP (e.g., results around PPL=60).

\end{document}